\documentclass{article}

\usepackage{arxiv}

\usepackage[utf8]{inputenc} % allow utf-8 input
\usepackage[T1]{fontenc}    % use 8-bit T1 fonts
\usepackage{hyperref}       % hyperlinks
\usepackage{url}            % simple URL typesetting
\usepackage{booktabs}       % professional-quality tables
\usepackage{nicefrac}       % compact symbols for 1/2, etc.
\usepackage{microtype}      % microtypography
\usepackage{lipsum}  
\usepackage{authblk}
\usepackage{graphicx}
\usepackage[square,sort,comma,numbers]{natbib}
\usepackage{doi}
\usepackage{xcolor}
\usepackage{framed}
\definecolor{shadecolor}{RGB}{252,252,252}
\usepackage{multirow}
\usepackage{subfigure}
\usepackage{caption}
\usepackage{amsmath}
\usepackage{cleveref}       % smart cross-referencing

\title{DARWIN Series: Domain Specific Large Language Models for Natural Science}

\date{August 24, 2023}
\author[1,2,*,**]{Tong Xie}
\author[2,3*]{Yuwei Wan}
\author[2,6]{Wei Huang}
\author[2,6]{Zhenyu Yin}
\author[2,6] {Yixuan Liu}
\author[1,2]{Shaozhou Wang}
\author[2,6]{Qingyuan Linghu}
\author[3]{Chunyu Kit}
\author[4,5]{Clara Grazian}
\author[6]{Wenjie Zhang}
\author[6]{Imran Razzak}
\author[1,**]{Bram Hoex}

\affil[1]{School of Photovoltaic and Renewable Energy Engineering, University of New South Wales, Kensington, NSW, Australia}
\affil[2]{GreenDynamics Pty. Ltd, Kensington, NSW, Australia}
\affil[3]{Department of Linguistics and Translation, City University of Hong Kong, Hong Kong, China}
\affil[4]{School of Mathematics and Statistics, University of Sydney, Camperdown, NSW, Australia}
\affil[5]{DARE ARC Training Centre in Data Analytics for Resources and Environments, Australia}
\affil[6]{School of Computer Science and Engineering, University of New South Wales, Kensington, NSW, Australia}
\affil[7]{School of Computing and Information Technology, University of Melbourne, Parkville, VIC, Australia}
\affil[*]{Authors contributed equally}
\affil[**]{Corresponding author: tong.xie@unsw.edu.au, b.hoex@unsw.edu.au}

\renewcommand{\headeright}{}

\hypersetup{
pdftitle={A template for the arxiv style},
pdfsubject={q-bio.NC, q-bio.QM},
pdfauthor={David S.~Hippocampus, Elias D.~Striatum},
pdfkeywords={First keyword, Second keyword, More},
}

\begin{document}
\maketitle

\begin{abstract}
Emerging tools bring forth fresh approaches to work, and the field of natural science is no different. In natural science, traditional manual, serial, and labour-intensive work is being augmented by automated, parallel, and iterative processes driven by artificial intelligence-based experimental automation and more. To add new capabilities in natural science, enabling the acceleration and enrichment of automation of the discovery process, we present DARWIN, a series of tailored LLMs for natural science, mainly in physics, chemistry, and material science. This series relies on open-source LLM, incorporating structured and unstructured scientific knowledge from public datasets and literature. We fine-tuned the models using over 60,000 instruction data points, emphasizing factual correctness. During the fine-tuning, we introduce the Scientific Instruction Generation (SIG) model, automating instruction generation from scientific texts. This eliminates the need for manual extraction or domain-specific knowledge graphs and efficiently injects scientific knowledge into the model. We also explore multi-task training strategies, revealing interconnections between scientific tasks. DARWIN series not only achieves state-of-the-art results on various scientific tasks but also diminishes reliance on closed-source AI models. Our research showcases the ability of LLM in the scientific domain, with the overarching goal of fostering prosperity within the broader AI for science community.
\end{abstract}

% keywords can be removed

\section{Introduction}
Code is the language of computers, and molecular atomic structure is the language of material science and chemistry \cite{Weston2019,Trewartha2022}. Discovering molecular atomic structures and their properties and utilizing them for practical applications is extremely time-consuming and labour-intensive. AI-driven approaches accelerate this by harnessing the material science knowledge developed over centuries through rigorous scientific procedures ensuring a cohesive and seamless framework. For example, AI can learn from large molecular database structures and then can be used to provide viable chemical compounds and their properties, which may accelerate the discovery of new materials or their development. 

\begin{figure}[h]
    \centering
    \includegraphics[width=16cm]{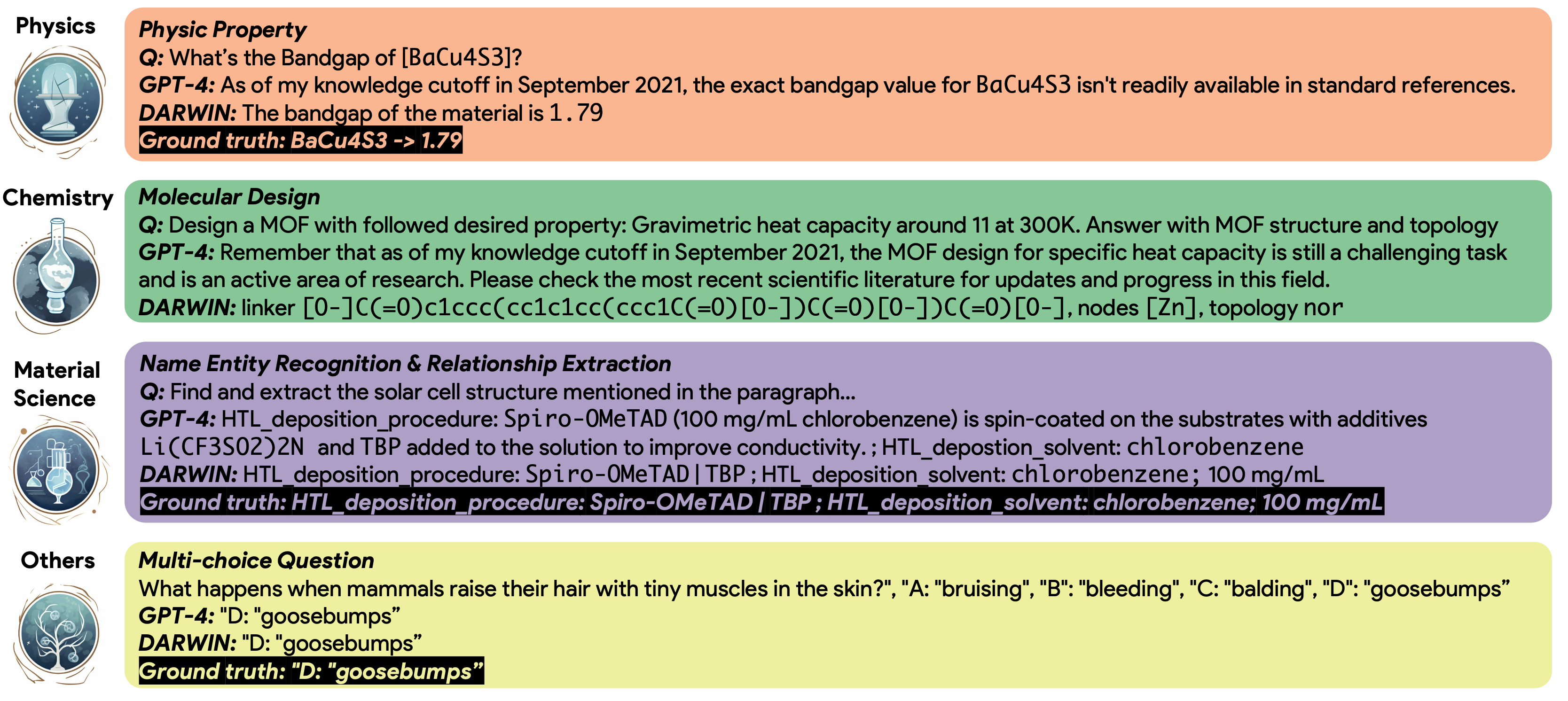}
    \caption{DARWIN vs GPT-4 comparative analysis in natural science tasks}
    \label{fig:enter-label}
    \vspace{-0.4cm}
\end{figure}

Instruction-tuned Large Language Models (LLMs), represented by InstructGPT \cite{Ouyang2022}, have captivated the academic community due to their proficiency in understanding directives and generating responses akin to human conversation. Compared to their smaller counterparts, LLMs demonstrate potent generalisation across various Natural Language Processing (NLP) tasks, illustrating their capacity to resolve unseen or intricate challenges in natural science. Even though fine-tuned GPT Models have shown an impressive ability to both domain-specific NLP tasks along with material and device prediction (MDP) \cite{xie2023large,Jablonka2023}, the core issue with these models lies in the fact that they remain inaccessible as open-source platforms, consequently compelling each user to engage in the laborious and financially burdensome task of individually fine-tuning the model on OpenAI's servers. Additionally, they fail to offer a robust degree of confidence in their precision, an aspect that can be viewed as a significant shortfall. Such constraints have the potential to impose a brake on the momentum at which LLMs are propelling scientific discovery.

The emergence of open-source alternatives, like LLaMA \cite{Touvron2023} and RWKV \cite{Peng2023}, has provided several methods to finetune foundational Large Language Models, like Alpaca \cite{alpaca} and Vicuna \cite{vicuna2023}. However, most of these instruction datasets are self-instructed by GPT-3.5 \cite{Ouyang2022} or GPT-4 \cite{openai2023gpt4}, which can lead to a reduced prediction accuracy \cite{Wang2022} (see Figure \ref{fig:enter-label}). This problem is particularly significant in natural science topics, as the above models are not explicitly designed for scientific fields such as materials science and chemistry. While there are examples in the medical field where knowledge graphs have been used to generate instruction data \cite{wang2023huatuo}, this is not common in natural science. Another costly approach involves GPT-3.5 or GPT-4 for multiple rounds of question-answering sessions with cancer-related papers. Despite these challenges, the evolution and improvement of LLM continue to open promising opportunities in various fields.

This study introduces DARWIN, a series of LLM meticulously tailored for the scientific domain, particularly for material science, chemistry, and physics. The models build upon the open-source LLaMA-7B \cite{Touvron2023} and achieve state-of-the-art results in various scientific tasks. In the fine-tuning process, we introduced the Scientific Instruction Generation (SIG) model, which automates the generation of instructions from scientific texts. This innovative approach streamlines the infusion of scientific knowledge into the model. By harnessing open-sourced scientific FAIR (Finable, Accessible, Interpreable, and Reuseable) datasets \cite{Wilkinson2016} and the scientific literature, we generated diverse instruction data, emphasising the factual correctness of the models' responses. The key contributions of this work are:

\begin{itemize}
    \item We present a novel LLM series named DARWIN that enhances performance across benchmarks in the scientific domain. This achievement represents a significant stride in our collective efforts to leverage LLMs for scientific inquiry, further endorsing their potential role in accelerating scientific discovery. Github: \url{https://github.com/MasterAI-EAM/Darwin}

    \item We introduce a cost-effective instruction generation approach based on high-quality content. By incorporating the academic papers into the instructions, the reliability of the training dataset is significantly enhanced. Furthermore, we employ the binding "To be continued" tokens for lengthy text input, circumventing the input token limitation inherent to LLaMA.

    \item We use open-source base models for training and public scientific benchmarks for evaluation, enforcing reproducibility in the long run. This open-source paradigm augments the accessibility of models and code and lowers the financial barriers for researchers, consequently fostering the development of the entire AI for Science community.    
\end{itemize}

\section{Related Work}

\subsection{Large language models}
The development of LLM has been an exciting journey, fueled by innovation and a constant push to improve capabilities while addressing critical concerns such as robustness, safety, and ethical implications. It began with simpler models such as Word2Vec \cite{mikolov2013distributed} and GloVe \cite{pennington2014glove}, which fundamentally changed how machines understood and processed natural language by enabling them to capture semantic and syntactic relationships between words. The advent of transformer models like BERT \cite{devlin2018bert} marked a key milestone in this journey, showcasing the power of attention mechanisms and the potential for pre-training on vast amounts of data. It set the stage for GPT, a series of models developed by OpenAI that pushed the boundaries of what LLMs could do, culminating in the groundbreaking GPT-3, a model with 175 billion parameters that showcased impressive capabilities in generating human-like text.

In the spirit of open research and a commitment to make 
LLMs task-adaptable, instruction-tuning \cite{Ouyang2022} is employed. The instructions are generated from human-annotated prompts and feedback \cite{Ouyang2022}, or public benchmarks and datasets with manual or automatic augmentation \cite{wang2022benchmarking}. Compared to the above approaches, self-instruct \cite{Wang2022} generates new tasks (instructions along with instances) from scratch and has been quickly adopted for its superior performance and low cost. For instance, Stanford Alpaca \cite{alpaca} is trained on 52k instructions generated by GPT-3.5, and Vicunna \cite{vicuna2023} uses around 700k instructions from conversations gathered from ShareGPT.com. There is a direct correlation between the size and diversity of the instruction data and the generalizability of resulting models to unseen tasks \cite{wang2022super,chung2022scaling}. While self-instruct tuning brings qualitative behaviours to LLMs, researchers find an effect of model collapse that the use of model-generated content in training may cause irreversible defects in the resulting models, where tails of the original content distribution disappear \cite{shumailov2023curse}. 

\subsection{Scientific pretrained models}

Before the era of billion-parameter models, there have been several research efforts to use scientific corpora for language modeling in natural science, such as SciBERT \cite{beltagy2019scibert} and PubMedBERT \cite{gu2021domain}, or in specific domains, such as MatBERT \cite{trewartha2022quantifying} and DNABERT \cite{ji2021dnabert}. Characterized by the larger computational clusters, expanded datasets, and increased model parameters, the advent of the GPT-3 \cite{Brown2020} has been a game-changer. This model has been adapted to tackle many NLP tasks in the form of question-answering, for example, merely prompting it with chemical queries in natural language. 

Some studies train LLMs tailored to specific scientific domains, exemplified by endeavors such as PMC-LLaMA \cite{wu2023pmc} and Med-PaLM \cite{singhal2022large} in the medical field. Within the broader natural science domain, Galactica \cite{taylor2022galactica} trained on a corpus encompassing 48 million examples of scientific articles, web content, textbooks, lecture notes, and encyclopedic references, claims to have multifaceted capability like solving mathematical problems and annotating molecular and protein structures. However, a fundamental problem with Galactica is that it is not able to distinguish truth from falsehood, which is normally called halluciation. In response to this challenge, some models have opted to integrate external APIs \cite{bran2023chemcrow} or knowledge graphs \cite{wang2023huatuo} to improve the reliability of generated content.

Simultaneously, scientific LLMs trained for specific tasks have demonstrated promising achievements. Notably, in the realm of biology, BioGPT \cite{luo2022biogpt} pre-trained on an extensive corpus of biomedical literature, has exhibited superior performance across various biomedical natural language processing (NLP) tasks, including relation extraction, document classification, and question-answering. This progress is further complemented by the field of chemistry. Jablonka et al. \cite{Jablonka2023} finetuned GPT-3 to solve tasks involving regression, classification, or inverse design and found finetuned GPT-3 often performs on par with, or even surpasses, conventional machine learning models while frequently requiring less training data. In addition, novel NLP tasks have been introduced to address domain-specific challenges. For instance, structured information inference (SII) \cite{xie2023large} was devised to tackle the information extraction and inference at the device level within the realm of materials science, and extracted information serves as instructive data for finetuning GPT-3, facilitating prediction of power conversion efficiency (PCE).

\section{Methods} 

In this section, we will introduce the training process of the series of DARWIN model, including datasets and modelling.

\subsection{Datasets}

Various science-related datasets were used in tuning to align LLM to the natural science domain. These generally include science exam questions answering dataset, retrieved questions and corresponding answers from scientific papers, and FAIR datasets.

\subsubsection{Scientific knowledge dataset}
\label{sub:sci}

\textbf{SciQ dataset.} 
Standardized tests drive progress in AI by challenging language understanding and pushing the boundaries of knowledge, modelling, and reasoning \cite{clark2015elementary}. The SciQ dataset \cite{welbl2017crowdsourcing} is a large crowdsourced science Questions \& Answers (QA) dataset comprising 13,679 science exam questions covering subjects like physics, chemistry, and biology. The questions are presented in multiple-choice format with four answer options, and most of them are accompanied by an additional paragraph containing supporting evidence for the correct answer. Since SciQ used science study textbooks as a base corpus, it provides broad and rather simple science knowledge.

\begin{figure}[h!]
\begin{minipage}[t]{0.48\linewidth}
  \centering
  \includegraphics[scale=0.2]{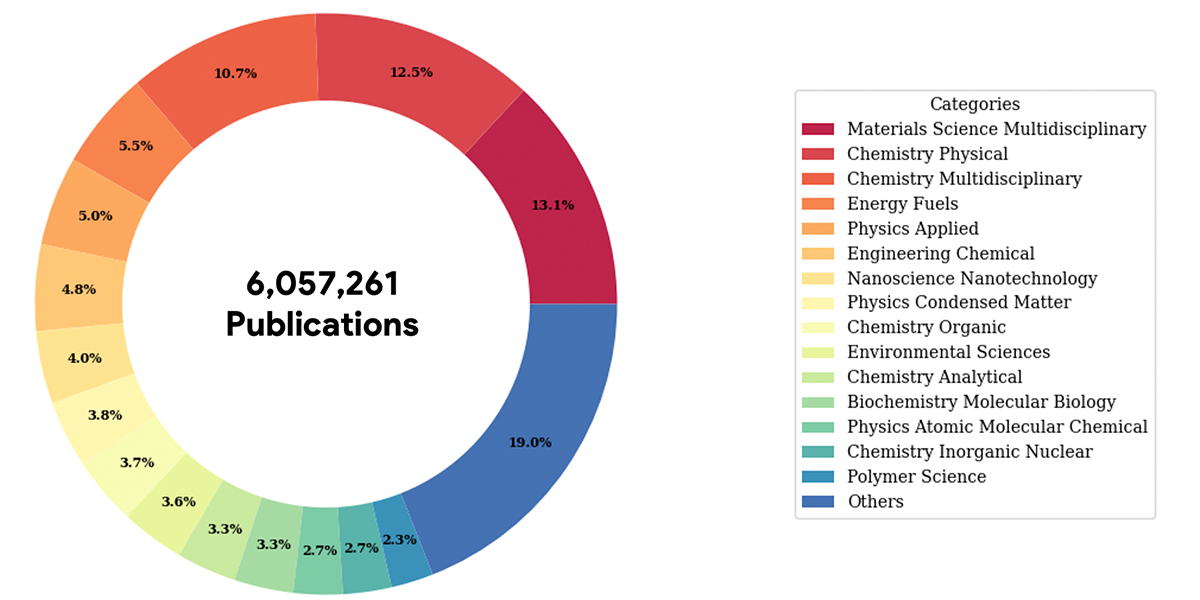}
  \caption{Composition of scientific paper dataset}
  \label{fig:sci-paper-cate}
\end{minipage}
  \hspace{0.2in}
\begin{minipage}[t]{0.48\linewidth}
\centering
  \includegraphics[scale=0.2]{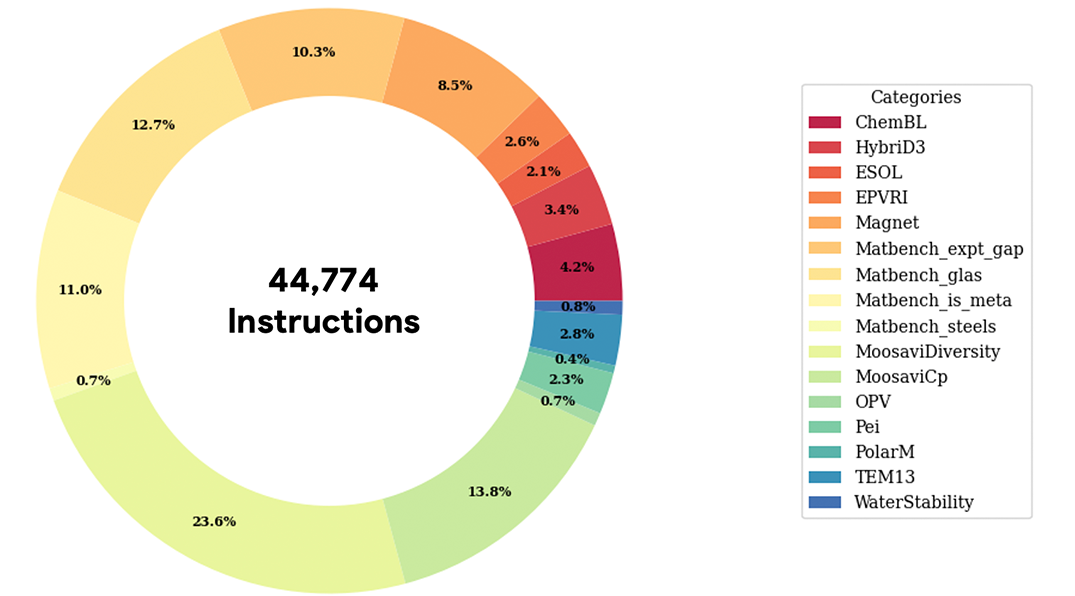}
  \caption{Composition of FAIR dataset}
  \label{fig:fair-cate}
\end{minipage}
\end{figure}

\noindent \textbf{Scientific paper datasets} 
\label{sub:sci_lit}
Utilizing the Web of Science, a collection of scholarly literature comprising 6 million papers was assembled, adhering to the following criteria: 1) categorized in the scientific disciplines of material science (see Figure \ref{fig:sci-paper-cate}), like chemistry, physics, energy, etc.; 2) available in full-text format from publishers such as Elsevier, Springer Nature, American Chemical Society, and Royal Society of Chemistry; and 3) written in the English language. We then proceeded to identify highly cited papers and utilized our web-scraping tool SciCrawler \footnote{https://github.com/MasterAI-EAM/SciCrawler} to download the content from different publishers and convert it into plain text. We finally employed cosine similarity of MatBert embeddings to remove papers with overly similar topics. This self-constructed paper dataset aims to enhance the model with professional science knowledge and facilitate significant progress in various domains.

\subsubsection{FAIR datasets}

FAIR stands for 'Findable, Accessible, Interoperable, and Reusable', which is a set of principles for enhancing the value and accessibility of data  \cite{Wilkinson2016}. Due to the strong impact of 4V (volume, variety, velocity, and veracity) of Big Data on materials science, efforts have been made in recent years to collect comprehensive data from research groups worldwide, including unpublished data, and ensure its FAIRness \cite{scheffler2022fair}. We collected 16 open-accessed FAIR datasets with multidisciplinary topics in physics, chemistry, and materials science, including but not limited to the physical structure, properties, synthesis and even material design (see Figure \ref{fig:fair-cate}. The details of these datasets and which tasks they are used to derive instructions can be found in Appendix Table~\ref{tb:fair}.

\subsection{Modeling}

For our model DARWIN, we developed a comprehensive training pipeline (see Figure~\ref{fig:pipeline}) utilizing different large language models, LLaMA-7B and Vicuna-7B models, to balance performance and cost. The whole training pipeline involves three LLMs: DARWIN-SIG, DARWIN-BASE and DARWIN-MDP.

\subsubsection{DARWIN-SIG}
\label{sub:sciqa}
% 说明SIG方式的意义【为什么不全文pretrain？有什么好处？为什么用vicunna？3.2.1写的步骤更清楚一点，GPT-4弄了几个？
Since pretraining on domain texts may not mitigate hallucination issues, we propose a cost-effective approach for generating instruction data using open-source model. The main idea is to train a generator for converting full-text scientific papers into question-answer pairs. By integrating these knowledge-based content into the instructions, we enhance the reliability of the training dataset and avoid model collapse induced by the utilization of training data fully generated by LLMs. The process of SIG can be break down as following steps:
\begin{enumerate}
    \item Prompt design: Through a process of iterative manual examination and seeking advice from domain experts, we were able to design an effective prompt for LLM to generate question-answer pairs with detailed information extracted from scientific papers.

\begin{figure*}[h!]
  \centering
  \includegraphics[width=0.8\textwidth]{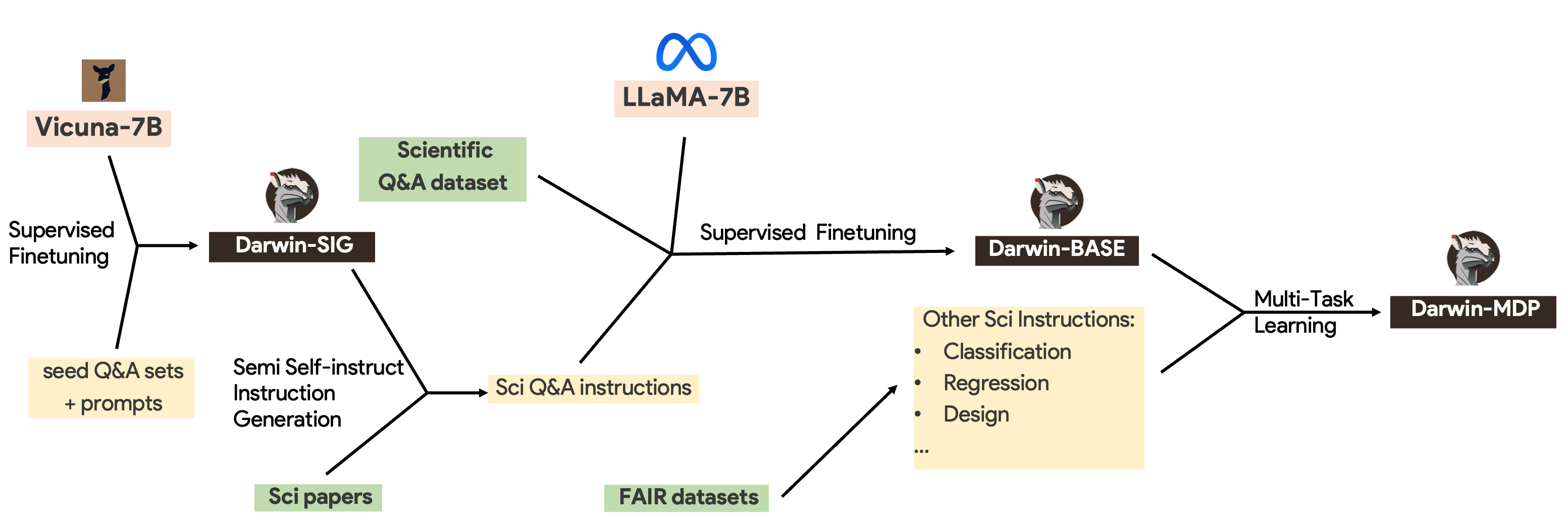}
  \caption{DARWIN Series Model Structure and Training Pipeline}
  \label{fig:pipeline}
\end{figure*}
    \item Seed QA generation: We select seed papers from the scientific paper dataset in Section \ref{sub:sci} and the rest are training papers. The seed papers were utilized as input to produce seed question-answer pairs by prompting GPT-4 (using prompt designed in step 1). 
    \item DARWIN-SIG: Our SIG model was developed by fine-tuning the LLM using seed papers and their corresponding question-answer pairs. Given a prompt $p$ and input text $x$, the goal of SIG is to generate a set of instructions $I$, each of which is a pair of question and corresponding answer $(q_i,a_i)$ related to $x$: $SIG(p,x) = I$. The initial prompt was converted into an instruction, while the seed papers served as input to the model. 
    \item Training QA generation: Utilizing the trained DARWIN-SIG, the large amount of training papers were subsequently used as input to automatically generate question-answer pairs with low cost. These QA pairs can be directly converted into instruction data.
\end{enumerate}
During the SIG process, there are some details that merit discussion. The first is the choice of base model. Considering the potential inference cost associated with GPT-3.5/4, we turn to use open-source model for training generator. And after human evaluation of performance, we chose Vicuna-7B model as the base model instead of LLaMA or Alpaca. The second is the length limit. The full-text papers are usually longer than limit of 2,048 tokens and for those lengthy ones, we segmented texts and continuously input them by using the ``[TBC]'' token as a connector (see Figure 5). Thus, the model could output question-answer pairs aligned with the complete content of the papers.

\begin{figure*}[h!]
  \centering
  \includegraphics[width=0.8\textwidth]{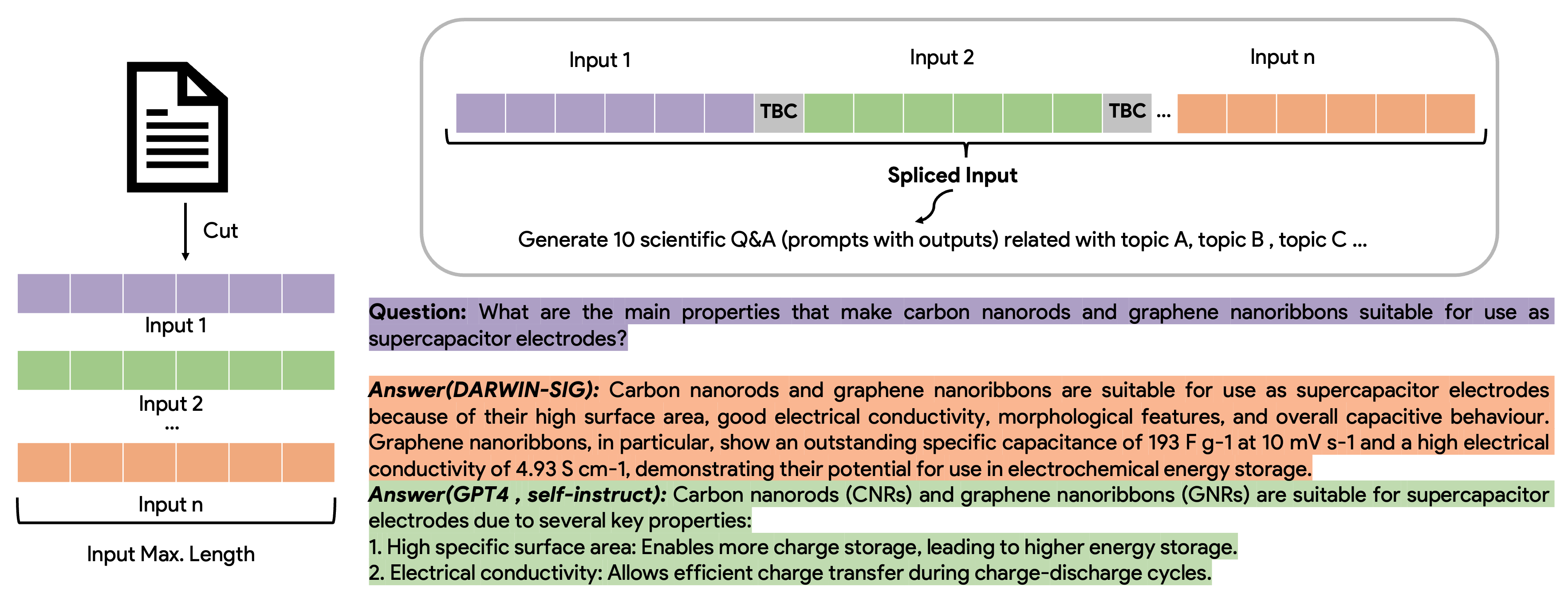}
  \caption{Darwin-SIG structure and comparison}
  \label{fig:pipeline}
\end{figure*}

We compare with self-questioning method \cite{fang2023mol} (see Appendix \ref{sec:sgi_self}). Self-questioning also generates factual question-answer pairs based on the given abstract by prompting GPT-3.5.it can be observed that question-answer pairs generated by DARWIN-SIG  are more informative and specific to the content of the input paper. Rather than being limited to simple definitions, the questions are complex enough to cover many aspects, including but not limited to the properties relevant to the application, the influence of strategies, structural differences, challenges, and experimental insights. Correspondingly, the answers generated are not overly generalized. Instead, they provided specific details, such as mentioning precise chemical elements and numerical values referenced in the original scientific papers. The high-quality Q\&A generation ensured the preservation of essential knowledge within lengthy scientific papers.

\subsubsection{DARWIN-BASE}
We generated knowledge-based instruction data for DARWIN based on SciQ dataset in section \ref{sub:sci} and generated question-answer pairs in \ref{sub:sciqa}. The answers in the SciQ dataset include correct options and provide support, which can be used as background knowledge input or as an explanation of the reasoning process. Based on this scenario, we generated a mixed set of instruction data consisting of the following patterns, arranged in increasing levels of difficulty: 1) Single-turn question-answering, where the model selects an answer based on the question and the provided background knowledge; 2) Single-turn question-answering, where the model selects an answer based on the question and also provides an explanation for the chosen answer; 3) Multi-turn dialogue, where the model selects an answer based on the question and provides an explanation for the chosen answer in response to user follow-up questions. Examples of the above three types are available in the Appendix (see Appendix \ref{sec: sciq_3}). Without any input, each question became an instruction, and its corresponding answer became an output. We fine-tuned the LLaMA-7B model using the two kinds of knowledge-based instruction data mentioned above, totalling over 20,000 instances. This process resulted in the development of DARWIN-BASE, a model equipped to engage in scientific question-answering. 

\subsubsection{DARWIN-MDP}
\label{subsub: boolean}

In addition to DARWIN which can conduct the scientific Q\&A task, we also involve 16 FAIR datasets (\ref{tb:fair}) to generate other instructions to further fine-tune DARWIN-BASE as DARWIN-MDP which can conduct the \textbf{M}aterial \& \textbf{D}evice \textbf{P}redictions tasks, such as classification, regression, and design etc.

\begin{itemize}

\item
\textbf{Classification}
\begin{snugshade*} \texttt{
\begin{quote}
    \textbf{"instruction":} "Tell me if given composition has glass-forming ability."\\
    \textbf{"input":}"Cr20Ni61P19" \\
    \textbf{"output":} "Yes, Cr20Ni61P19 has glass-forming ability." \\
\end{quote}
}
\end{snugshade*}
The classification task of scientific language models involves the categorization or labelling of scientific text data into predefined classes or categories. 
For example, DARWIN-MDP can be trained to classify chemical compounds based on their properties, such as solubility, toxicity, or stability. This can assist in drug discovery, material science, or chemical engineering applications.
An instance of the instruction of classification is shown above.

\item
\textbf{Regression}
\begin{snugshade*} \texttt{
\begin{quote}
    \textbf{"instruction":} "Write aqueous solubility expressed as a logarithm in mol/L of given SMILES in 25 ℃."\\
    \textbf{"input":}"CN(C)C(=O)Nc1ccc
    (Cl)c(Cl)c1" \\
    \textbf{"output":} "-3.8"\\
\end{quote}
}
\end{snugshade*}
A more demanding task than classification involves developing a regression model capable of predicting continuous property values, like open-circuit voltage (Voc), short-circuit current density (Jsc), fill factor (FF), and PCE etc. Although LLMs are unable to predict real numbers in regression tasks with high precision, they can still produce predictions of acceptable accuracy by employing rounded values during training. For device performance data, a precision of two decimal points is deemed sufficient. An example of the instruction of regression is shown above.

\item
\textbf{Inverse Design}
\begin{snugshade*} \texttt{
\begin{quote}
    \textbf{"instruction":} "Design a donor with acceptor PC61BM, PCE=0 for a organic solar cell with specific requirement"\\
    \textbf{"input":}"under AM1.5 and 100 light intensity" \\
    \textbf{"output":} "COC(=O)c1sc2csc(c3sc4c(c3F) c(OC)c3scc(F)c3c4OC)c2c1F" \\
\end{quote}
}
\end{snugshade*}

In the context of material science, inverse design refers to a computational approach that aims to design new materials or optimize existing ones by working backward from desired properties or functionalities \cite{Jablonka2023}. Instead of starting with a specific material and analyzing its properties, inverse design begins with a desired set of properties or performance targets and seeks to identify the material structure or composition that would exhibit those properties. By utilizing inverse design techniques, researchers can potentially discover novel materials with tailored properties that are challenging or time-consuming through traditional experimental approaches. The inverse design has shown promise in various areas of material science, including photovoltaics, catalysts, batteries, thermoelectric materials, and more. An example of the instruction of inverse design is shown above.

\end{itemize}

\section{Experiments}

\subsection{Baselines} 

To demonstrate the performance of DARWIN on different tasks, we compared the performance of DARWIN with 2 baseline models.
\begin{itemize}
\item
\textbf{LLaMA}  is an LLM released by Meta AI with a variety of model sizes from 7 billion to 65 billion parameters. LLaMA-7B version is used as our base model out of its ease of training.
\item
\textbf{GPT-3} is the third-generation language model with 175 billion parameters created by OpenAI. 
\end{itemize}
In addition to the aforementioned two baselines, we included results from several dedicated machine learning models as references, like MODNet (v0.1.1), MolCLR and AMMExpress v2020. Those results were from matbench \cite{Dunn2020}. It should be noted that each model was individually trained on a specific FAIR dataset. 

\subsection{Fine-tuning procedure}

We conducted fine-tuning of the LLaMA-7B model with instruction tuning on an 8-A100 GPU. The hyperparameters we used were similar to the ones used with Alpaca: the fine-tuning procedure included training for 3 epochs with a batch size of 4 for both training and evaluation. Gradients were accumulated over 8 steps, the learning rate was set to 2e-5 with no weight decay, and a warm-up ratio of 0.03. We utilized the Brain Floating Point 16 (BF16) data format to strike a balance between precision and computational cost. We fine-tuned LLaMA-7B on subsets or combinations of instruction data and named them differently as follows:
\begin{itemize}
\item 
\textbf{gptchem} was fine-tuned on several FAIR datasets separately, using GPT-3 as base model \cite{jablonka2023gpt}.  
\item
\textbf{LLaMA-single} was fine-tuned on instruction data from each dataset of 12 FAIR datasets, which means there are 12 LLaMA-single.
\item
\textbf{LLaMA-all} was fine-tuned on mixed instruction data of 9 FAIR datasets.
\item
\textbf{DARWIN} was fine-tuned on instruction data from SciQ dataset and question-answer pairs generated by SIG.
\item
\textbf{DARWIN-MDP} was further fine-tuned on DARWIN on mixed instruction data of 16 FAIR datasets.
\end{itemize}

Our final model DARWIN-MDP is a language model fine-tuned using supervised learning from a LLaMA-7B model on over 60,000 instruction-following demonstrations derived from scientific literature and datasets.

\subsection{Metrics}

We use accuracy of multiple choice of SciQ as metric to evaluate performance in fine-tuning on scientific knowledge. If the model can choose the correct option and generate the correct corresponding answer, then it is counted as correct: 
\begin{equation*}
\text{{Accuracy}} = \frac{{\text{{\# of correctly answered questions}}}}{{\text{{total \# of questions}}}}.
\end{equation*}
We also evaluated the performance of the fine-tuned DARWIN model on each MDP task separately. We only evaluate classification and regression task, but leave design task for future work due to its difficulty of evaluation.

For the classification task, we evaluted the performance of the model on each dataset using F1 score. When dealing with boolean-type labels (as exemplified in Section \ref{subsub: boolean}), we compare the predictions, expressed as 'Yes' or 'No', with the ground truth values ('True' or 'False'), and evaluate the F1-binary metric specifically for the class indicated by pos\_label. For multi-category labels, we use macro-F1 score (unweighted mean of the F1 scores calculated per class). 
% \begin{align*}
% \text{{Precision}} &= \frac{{\text{{True Positives}}}}{{\text{{True Positives}} + \text{{False Positives}}}} \\
% \vspace{0.02cm} \\
% \text{{Recall}} &= \frac{{\text{{True Positives}}}}{{\text{{True Positives}} + \text{{False Negatives}}}} \\
% \vspace{0.02cm} \\
% \text{{F1 score}} &= 2 \times \frac{{\text{{Precision}} \times \text{{Recall}}}}{{\text{{Precision}} + \text{{Recall}}}}
% \end{align*}

For the regression task, we use Mean Absolute Error (MAE) to measure the difference between the predicted value and the true value:  
\begin{equation*}
\text{{MAE}} = \frac{1}{n} \sum_{i=1}^{n} |y_i - \hat{y}_i|.
\end{equation*}

\section{Results}

In this section, we present the performance of the DARWIN-BASE model and the DARWIN-MDP model, comparing them with the existing metrics from a previous study. 

\subsection{DARWIN-BASE Model Performance}

Table \ref{table:1} presents a comparative analysis of zero-shot and training outcomes concerning the SciQ dataset. SciQ serves as a widely recognized benchmark for scientific question-answering tasks, and numerous large language models (LLMs) employ it for assessing their zero-shot performance. Notably, the metrics reported in previous studies predominantly reflect zero-shot results, which tend to exhibit relatively lower accuracy.

For our evaluation, we conducted a random selection of 1000 questions from the SciQ dataset to serve as a test set. Remarkably, the performance of the DARWIN model on the SciQ dataset demonstrates an impressive accuracy rate of 96.9\%, despite having a model size of only 7 billion parameters. It is crucial to emphasize that these results were obtained from previously unseen data, despite DARWIN's fine-tuning on a subset of the SciQ dataset. This underscores the exceptional performance of DARWIN, particularly when juxtaposed with models that lack domain-specific fine-tuning.

\begin{table}[h!]
\centering
    \caption{Comparative evaluation of DARWIN and SOTA models on SciQ dataset}
    \begin{tabular}{ccc}
    \toprule
    \textbf{Number of Examples} & \textbf{Model} & \textbf{Accuracy}\\
    \midrule
    \multirow{3}*{zero-shot} & BLOOM (7 B) &  89.2\\
            & OPT (6.7 B) &   90.1\\
            & Pythia (Deduplicated) (6.9 B) &   91.1\\
    \midrule
    \multirow{1}*{five-shot} & Pythia (Deduplicated) (6.9 B) &   95.2\\
    \midrule
    \multirow{1}*{training} & DARWIN-BASE (7 B)  & 96.9\\
  \bottomrule
\end{tabular}
\label{table:1}
\end{table}

%2. mdp各任务的表

\subsection{DARWIN-MDP Model Performance}

Table \ref{table:2} and Table \ref{table:3} present the performance metrics of the DARWIN-MDP model across various regression and classification tasks. Notably, DARWIN-MDP consistently achieves the lowest Mean Absolute Error (MAE) in most regression tasks and the highest F1 score in most classification tasks. Comparing LLaMA-single and LLaMA-all, we observed that fine-tuning on multiple tasks together (LLaMA-all) yields superior performance compared to fine-tuning on individual tasks separately (LLaMA-single). This highlights the effectiveness of multi-task learning in enhancing the model's overall capabilities. Furthermore, when comparing LLaMA-all with DARWIN-MDP, we find that the initial stage of fine-tuning scientific knowledge significantly contributes to further improvements in performance. This underscores the importance of domain-specific pre-training, as DARWIN-MDP, with its initial training on scientific knowledge, outperforms the LLaMA-all model. Finally, the performance of DARWIN-MDP is comparable with the dedicated machine learning model for specific tasks, which provides a novel material design method, one model could finish multiple scientific tasks with simple prompting.

Overall, the results demonstrate that both the DARWIN series exhibit exceptional performance, surpassing the existing metrics reported in the previous study. These findings validate the effectiveness of domain-specific fine-tuning and multi-task learning, reinforcing the potential of these models for a wide range of natural language processing and scientific tasks.

\begin{table*}[!tbh]
\centering
\caption{
Comparative evaluation of DARWIN and SOTA models for Regression Task (in MAE).
}
\begin{tabular}{c|cccccccc}
\hline
\textbf{Model} & \textbf{OPV} & \textbf{steels} & \textbf{ChEML} & \textbf{expt\_gap} & \textbf{ESOL} & \textbf{Naga}  & \textbf{Diversity} & \textbf{Cp} \\
\hline
MODNet (v0.1.12) & / & \textbf{88} & / & 0.33 & / & / & / \\
MolCLR & / & / & 0.6706 & / & 0.9345 & / & /\\
gptchem & 0.65 & 142 & 0.9284 & 0.46 & 1.1544 & \textbf{1.97} & / & /\\
LLaMA-single & 0.56 & 229.88 & 0.907 & 0.406 & 0.775 & 2.52 & 0.38 & 4.89\\
LLaMA-all & 0.56 & 160.81 & 0.81 & \textbf{0.31} & 0.74 & / & / & /\\
DARWIN-MDP & \textbf{0.54} & 125.97 & \textbf{0.798} & 0.317 & \textbf{0.71} & 2.38 & \textbf{0.35} & \textbf{4.75}\\
\hline
\end{tabular}
\label{table:2}
\end{table*}

\begin{table*}[!htb]
\centering
\caption{
Comparative Evaluation of DARWIN and SOTA models for Classification Task (in F1 Score).
}
\begin{tabular}{c|cccc}
\hline
\textbf{Model} & \textbf{Pei} & \textbf{waterStability} & \textbf{matbench\_is\_metal} & \textbf{matbench\_glass}  \\
\hline
MODNet (v0.1.12) & / & / & 91.5 & \textbf{97.5} \\
AMMExpress v2020 & / & / & 92.0 & 90.4\\
gptchem & 79.2 & 74.4 & 89.0 & 88.0 \\
LLaMA-single & 90.8 & 71.0 & 89.1 & 86.7 \\
LLaMA-all & 89.1 & 86.7 & \textbf{96.5} & 87.7 \\
DARWIN-MDP & \textbf{97.2} & \textbf{93.3} & 95.9 & 88.0 \\
\hline
\end{tabular}
\vspace{-0.5cm}
\label{table:3}
\end{table*}

\section{Conclusion}

The emergence of foundational AI models and expansive language models offers the remarkable potential to redefine the advancement of scientific AI, making it more accessible, accurate, and cross-domain. However, the scientific domain, with its inherent complexities, presents unique challenges when it comes to implementing large language models.\\
Our research offers a glimpse into the possibilities and hurdles of employing these technologies in scientific applications. In this study, we incorporated existing knowledge bases effectively into LLM and created the DARWIN series which achieved SOTA results on multiple scientific tasks. We also introduced a SIG model that can automatically convert knowledge from a large corpus of papers. Furthermore, we explored different multi-task training strategies and found that training on knowledge-based questions and answers first can improve prediction performance on FAIR datasets, and combining multiple tasks proved to be more effective than individual training, highlighting the underlying connections between seemingly independent tasks.
\vspace{-0.15cm}
\section{Limitations}

\textbf{Data:} A key limitation of this study stems from the limited quantity of question-answer pairs and dependence on the somewhat confined FAIR datasets, potentially inhibiting the model's applicability and effectiveness. To address this, future research could focus on amplifying the volume of tasks, generating more question-answer pairs via SIG, thereby enhancing dataset representativeness and improving model performance across various domains.

\textbf{Model:} In striving for desired performance, this study utilizes multiple models instead of a unified one, introducing complexity and potential inefficiencies. To overcome this, future research should focus on scaling up the model using efficient tuning methods for comparable or superior performance with limited resources. Also, the impact of fine-tuning on large-scale scientific text as a means of injecting knowledge into the model remains untested. Future work will address this through comparative analysis, providing valuable insights into the model's ability to leverage additional knowledge sources and improve performance in scientific fields.

\textbf{Evaluation:} DARWIN-BASE is employed as the foundational model, with its performance evaluated in terms of multiple-choice question answering. A comprehensive evaluation mechanism for the question-answering task is warranted to appraise the efficacy of knowledge injection from diverse perspectives. We intend to formulate a model-graded evaluation mechanism that segregates answers into various categories contingent on their quality and compliance with specific criteria, such as misunderstanding, relevance, or correctness. This methodology would afford a more exhaustive assessment of the model's performance, obviating the need for extensive manual intervention.

%\section*{Acknowledgements}
% This document has been adapted by Yue Zhang, Ryan Cotterell and Lea Frermann from the style files used for earlier ACL and NAACL proceedings, including those for 
% ACL 2020 by Steven Bethard, Ryan Cotterell and Rui Yan,
% ACL 2019 by Douwe Kiela and Ivan Vuli\'{c},
% NAACL 2019 by Stephanie Lukin and Alla Roskovskaya, 
% ACL 2018 by Shay Cohen, Kevin Gimpel, and Wei Lu, 
% NAACL 2018 by Margaret Mitchell and Stephanie Lukin,
% Bib\TeX{} suggestions for (NA)ACL 2017/2018 from Jason Eisner,
% ACL 2017 by Dan Gildea and Min-Yen Kan, NAACL 2017 by Margaret Mitchell, 
% ACL 2012 by Maggie Li and Michael White, 
% ACL 2010 by Jing-Shin Chang and Philipp Koehn, 
% ACL 2008 by Johanna D. Moore, Simone Teufel, James Allan, and Sadaoki Furui, 
% ACL 2005 by Hwee Tou Ng and Kemal Oflazer, 
% ACL 2002 by Eugene Charniak and Dekang Lin, 
% and earlier ACL and EACL formats written by several people, including
% John Chen, Henry S. Thompson and Donald Walker.
% Additional elements were taken from the formatting instructions of the \emph{International Joint Conference on Artificial Intelligence} and the \emph{Conference on Computer Vision and Pattern Recognition}.

%% Define the bibliography file to be used
%\bibliography{references}
\bibliographystyle{unsrt}
\bibliography{template}  %%% Uncomment this line and comment out the ``thebibliography'' section below to use the external .bib file (using bibtex) .

\clearpage
\appendix
\onecolumn
\section{Categories of collected papers}
\begin{table*}[!htb]
\centering
\begin{tabular}{c|cccc}
\hline
Web of Science Categories & Record Count & Ratio \\ 
\hline	
Materials Science Multidisciplinary & 969264 & 16.002\\
Chemistry Physical	& 919588 & 15.182\\
Chemistry Multidisciplinary & 788218 & 13.013\\
Energy Fuels & 403050 & 6.654\\
Physics Applied & 366015 & 6.043\\
Engineering Chemical & 355343 & 5.866\\
Nanoscience Nanotechnology & 298217 & 4.923\\
Physics Condensed Matter & 277983 & 4.589\\
Chemistry Organic & 274913 & 4.539\\
Environmental Sciences & 262377 & 4.332\\
Chemistry Analytical & 246136 & 4.063\\
Biochemistry Molecular Biology & 242381 & 4.001\\
Physics Atomic Molecular Chemical & 201022 & 3.319\\
Chemistry Inorganic Nuclear & 199876 & 3.300\\
Polymer Science & 173356 & 2.862\\
Metallurgy Metallurgical Engineering & 170305 & 2.812\\
Chemistry Applied & 162747 & 2.687\\
Electrochemistry & 153127 & 2.528\\
Engineering Environmental & 148581 & 2.453\\
Physics Particles Fields & 137267 & 2.266\\
Thermodynamics & 135296 & 2.234\\
Mechanics & 133424 & 2.203\\
Nuclear Science Technology & 123304 & 2.036\\
Chemistry Medicinal & 123152 & 2.033\\
Food Science Technology & 118760 & 1.961\\
\hline
\end{tabular}
\end{table*}
Showing 25 out of 245 entries.
\section{Comparison of Self-questioning and SIG}
\label{sec:sgi_self}
Self-questioning:\\
\textbf{Prompt}: \\
Given a paragraph of paper, can you ask five questions and give answers that can be directed to the paragraph? List in the format of Question-Answer pair.\\
Paragraph: One- and two-dimensional carbon nanomaterials are attracting considerable attention because of their extraordinary electrical, mechanical and thermal properties, which could lead to a range of important potential applications. Synthetic processes associated with making these materials can be quite complex and also consume large amounts of energy, so a major challenge is to develop simple and efficient methods to produce them. Here, we present a self-templated, catalyst-free strategy for the synthesis of one-dimensional carbon nanorods by morphology-preserved thermal transformation of rod-shaped metal–organic frameworks. The as-synthesized non-hollow (solid) carbon nanorods can be transformed into two- to six-layered graphene nanoribbons through sonochemical treatment followed by chemical activation. The performance of these metal–organic framework-derived carbon nanorods and graphene nanoribbons in supercapacitor electrodes demonstrates that this synthetic approach can produce functionally useful materials. Moreover, this approach is readily scalable and could be used to produce carbon nanorods and graphene nanoribbons on industrial levels.\\
\textbf{GPT-3.5 completion}: \\
Q1: What is being discussed in this paragraph? \\
A1: This paragraph is discussing the synthesis of one-dimensional carbon nanorods through a self-templated, catalyst-free strategy and the potential application of these nanorods and graphene nanoribbons in supercapacitor electrodes. \\
\\
Q2: What transforms the nanorods into graphene nanoribbons? \\
A2: Sonochemical treatment followed by chemical activation transforms the nanorods into graphene nanoribbons.\\
\\
Q3: What properties of nanomaterials makes them attractive? \\
A3: The extraordinary electrical, mechanical and thermal properties of one- and two-dimensional carbon nanomaterials makes them attractive. \\
\\
Q4: What is the major challenge associated with the synthesis of these materials? \\
A4: The major challenge in the synthesis of these materials is to develop simple and efficient methods that consume low amounts of energy.  \\
\\
Q5: What advantages does this particular synthetic approach offer?\\
A5: This particular synthetic approach offers advantages such as scalability and the production of functionally useful materials on industrial levels.\\
%\rule[-10pt]{16cm}{0.05em}
\hrule
~\\
SIG:\\
\textbf{Prompt:}\\ Please generate 10 scientific Q\&A (prompts with outputs) related with "six-layered graphene nanoribbons", "one-dimensional carbon nanorods", "fig", "mof-74", "mof-74-rod", "synthesis", "carbon nanotubes", "excellent capacitive behaviour", "formation", "materials", "mpc", "acid", "chemical activation", "mofs", "useful materials". Don't ask very simple questions, like definition questions (e.g. What is XXX). You should generate more complex problems. Answer using the data from provided information. Add detail to answers as much as possible, such as answer the specific chemical elements and numbers. (input is the full text of paper)\\
\textbf{SIG Completion:}\\
Q1: What are some potential applications of six-layered graphene nanoribbons synthesized by chemical activation of a defective graphene layer?\\
A1: Six-layered graphene nanoribbons have a high surface area, which makes them useful for gas storage and catalyst support. They can also be used in electrochemical sensing and separation, and as a special template.\\
\\
Q2: How does the chemical activation process work in the synthesis of six-layered graphene nanoribbons?\\
A2: The chemical activation process involves treating a graphene layer with a strong acid, such as sulfuric acid or nitric acid, in a vacuum atmosphere. The acid attacks the weak interlayer dangling bonds, causing them to break and forming strong covalent bonds between the layers.\\
\\
Q3: What is the significance of the fig graphite intercalation behavior in six-layered graphene nanoribbons?\\
A3: The fig graphite intercalation behavior in six-layered graphene nanoribbons indicates that there are still many voids left between the layers after intercalation with KCl. This is because the intercalation process does not fill all of the defects in the graphene layer.\\
\\
Q4: What are the key features of the MOF-74 compound and its role in six-layered graphene nanoribbons?\\
A4: MOF-74 is a metal-organic framework that exhibits highly connected interior cavities. It is used as a template to create the six-layered graphene nanoribbons, in which the MOF-74 molecules act as glue to hold the graphene layers together.\\
\\
Q5: What are the key steps in the synthesis of one-dimensional carbon nanorods?\\
A5: The synthesis of one-dimensional carbon nanorods involves depositing carbon onto a substrate in the form of rods, heating the substrate, and then removing the substrate. This process can be carried out using a variety of substrates, such as silicon, ceramic, metal, or glass.\\
\\
Q6: What chemical activation methods can be used for the synthesis of six-layered graphene nanoribbons?\\
A6: Two common chemical activation methods for the synthesis of six-layered graphene nanoribbons are treatment with sulfuric acid (H2SO4) and nitric acid (HNO3).\\
\\
Q7: What are the main defects present in the graphene layer after chemical activation for the synthesis of six-layered nanoribbons?\\
A7: The main defects present in the graphene layer after chemical activation for the synthesis of six-layered nanoribbons are disconnected regions due to the separation of carbon layers from the graphene sheet, as well as chains of defects running along the graphene sheets.\\
\\
Q8: What are the key features of the MOF-74 compound used in the six-layered graphene nanoribbon synthesis method?\\
A8: MOF-74 is a metal-organic framework consisting of 74 atoms, 48 of which are carbon. It has large pore sizes, with a average diameter of around 10 Å, and high surface area, with a BET surface area of around 700 m2/g.\\
\\
Q9: How do the electrochemical properties of MOF-74-based materials compare to those of graphene nanoribbons?\\
A9: MOF-74-based materials have a larger specific surface area, higher oxygen storage capacity, and better electrochemical stability than graphene nanoribbons. They also exhibit superior thermal stability, with melting temperatures higher than 1,000 degrees.\\
\\
Q10: What are the main factors that contribute to the excellent capacitive behavior of MOF-74-based materials?\\
A10: The main factors that contribute to the excellent capacitive behavior of MOF-74-based materials are the large specific surface area, high oxygen storage capacity, and the presence of defects in the graphene layer, which lead to a large number of trapped charge carriers.

\section{Three patterns of SciQ instruction data}
\label{sec: sciq_3}
\textbf{1) Open-book single-turn question-answering}\\
instruction: Read the following paragraph and choose an answer for a multiple choice question about the paragraph\\ 
input: Oxidants and Reductants Compounds that are capable of accepting electrons, such as O2 or F2, are calledoxidants (or oxidizing agents) because they can oxidize other compounds. In the process of accepting electrons, an oxidant is reduced. Compounds that are capable of donating electrons, such as sodium metal or cyclohexane (C6H12), are calledreductants (or reducing agents) because they can cause the reduction of another compound. In the process of donating electrons, a reductant is oxidized. These relationships are summarized in Equation 3.30: Equation 3.30 Saylor URL: http://www. saylor. org/books.\textbackslash n Question: Compounds that are capable of accepting electrons, such as o 2 or f2, are called what? (A) residues (B) antioxidants (C) Oxygen (D) oxidants\\
output: (D) oxidants.\\
\textbf{2) Closed-book single-turn question-answering}\\
instruction: Choose an answer for this multiple choice question and explain: Compounds that are capable of accepting electrons, such as o2 or f2, are called what? (A) residues (B) antioxidants (C) Oxygen (D) oxidants\\
input: Oxidants and Reductants Compounds that are capable of accepting electrons, such as O2 or F2, are calledoxidants (or oxidizing agents) because they can oxidize other compounds. In the process of accepting electrons, an oxidant is reduced. Compounds that are capable of donating electrons, such as sodium metal or cyclohexane (C6H12), are calledreductants (or reducing agents) because they can cause the reduction of another compound. In the process of donating electrons, a reductant is oxidized. These relationships are summarized in Equation 3.30: Equation 3.30 Saylor URL: http://www. saylor. org/books.\
output: (D) oxidants.\\
\textbf{3) Multi-turn dialogue}\\
instruction: <user>: Choose an answer for this multiple choice question: Compounds that are capable of accepting electrons, such as o2 or f2, are called what? (A) residues (B) antioxidants (C) Oxygen (D) oxidants <bot>: (D) oxidants. <user>: Can you explain why?\\
input:  \\
output: Oxidants and Reductants Compounds that are capable of accepting electrons, such as O2 or F2, are calledoxidants (or oxidizing agents) because they can oxidize other compounds. In the process of accepting electrons, an oxidant is reduced. Compounds that are capable of donating electrons, such as sodium metal or cyclohexane (C6H12), are calledreductants (or reducing agents) because they can cause the reduction of another compound. In the process of donating electrons, a reductant is oxidized. These relationships are summarized in Equation 3.30: Equation 3.30 Saylor URL: http://www. saylor. org/books.
\section{MDP regression task}
\textbf{OPV}: \\
instruction: what is the power conversion efficiency of organic solar cells with the donor: COC(=O)c1sc2csc(-c3sc4c(c3F)c(OC)c3scc(F)c3c4OC)c2c1F, and Acceptors: PC61BM? \\
Output: 0.96478299093449994 \\
\textbf{Matbench\_steels}: \\
instruction: Write a possible yield strength of given composition at 800-1200 \u00b0C. \\
input: Fe0.721C0.000476Mn0.000104Si0.000203Cr0.0560Ni0.109Mo0.0304V0.\\0105Nb0.0000615Co0.0611Al0.00191Ti0.00931 \\
output: 1040.4 \\
\textbf{ChEML}: \\
instruction: Write lipophilicity of given SMILES. \\
input: CC(C)C(NC(=O)CN1C(=O)C(=CN=C1C2CCCCC2)NC(=O)OCc3ccccc3)C(=O)C(F)(F)F \\
output: 2.87 \\
\textbf{Matbench\_expt\_gap}:
instruction: What is band gap of given composition? \\
input: Sr5Cd2Sb5F \\
output: 0.0 \\
\textbf{ESOL}: \\
instruction: What is water solubility of given SMILES in room temperature?\\
input: CCOP(=S)(OCC)SCSCC \\
output: 7.762471166286911e-05 \\ 
\begin{figure*}[h!]
  \centering
  \includegraphics[width=0.7\textwidth]{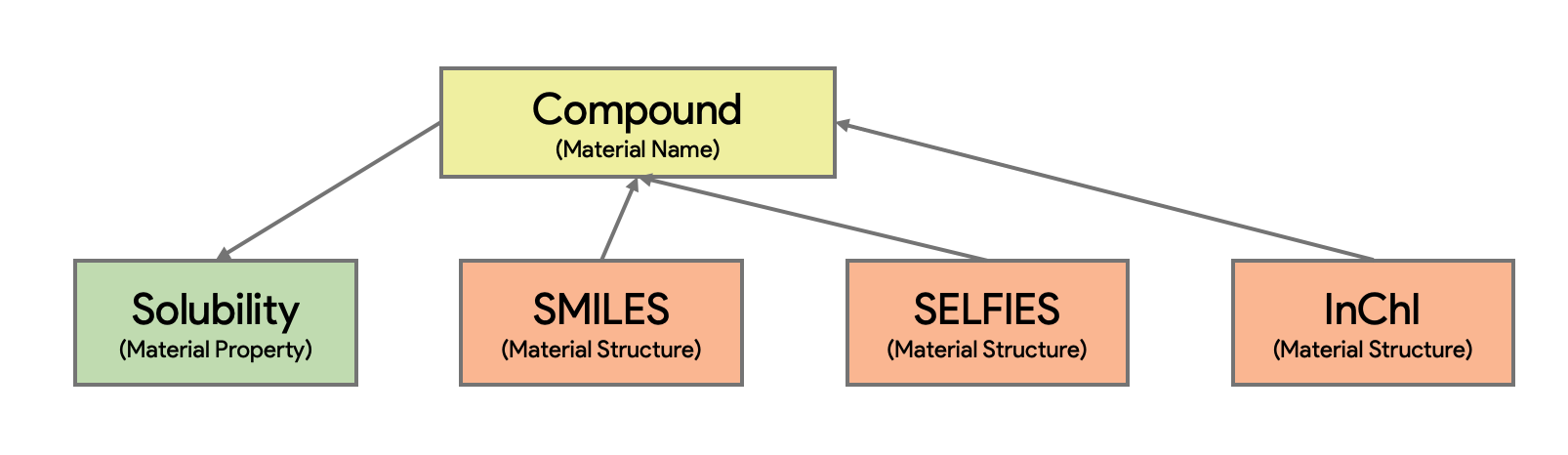}
  \caption{ESOL task instruction generation structure,
A solid line delineates a one-to-one correspondence}
  \label{fig:pipeline}
\end{figure*} \\
\textbf{MoosaviCp}: \\
instruction: What is weight-averaged molecular weight of given SMILES at normal measurement conditions.? \\
input: CC1=CC(N(CC(CCCCCC)CCCC)C(C2=C3SC(C4=CC5=C(C(SC(C)=C6)=C6[Si]5\\(CCCCCCCC)CCCCCCCC)S4)=C2)=O)=C3S1 \\ 
output: 30.5 \\
\begin{figure*}[h!]
  \centering
  \includegraphics[width=0.7\textwidth]{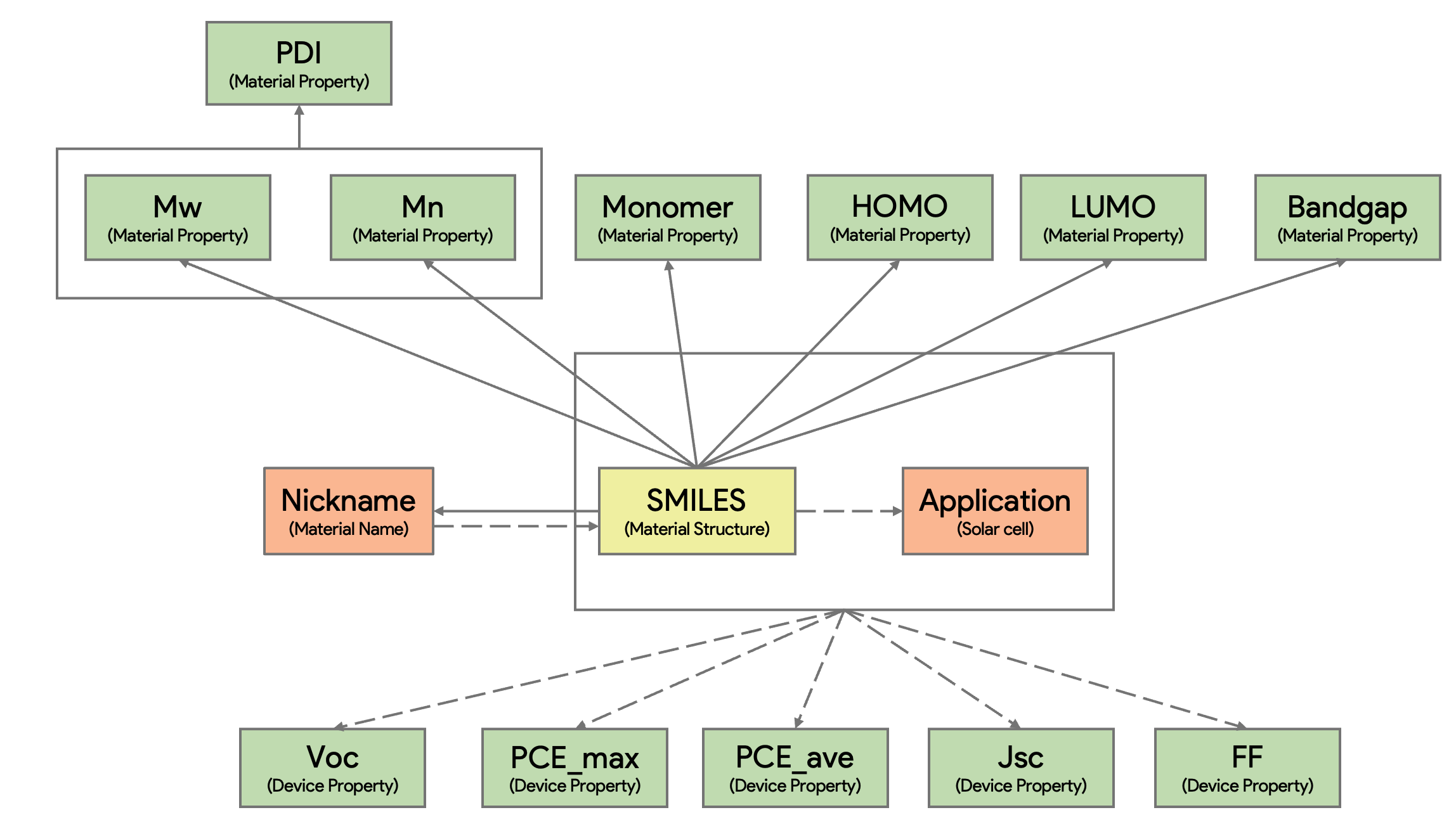}
  \caption{OPV task instruction generation structure,
A solid line delineates a one-to-one correspondence, while a dashed line symbolizes a one-to-many association.}
  \label{fig:pipeline}
\end{figure*} \\
\textbf{MoosaviDiversity}: \\
instruction: What is CellV for given MOF? \\
input: AVIVAC\_clean \\
output: 18049.1 \\
\textbf{MoosaviCp}: \\
instruction: Write the predicted gravimetric heat capacity at 350 K of given MOF with MOF structural features and topology. \\
input: linker OC(=O)C1=CC(=[C]C(=C1)C(=O)O)Oc1cccc(c1C(=O)[O-])C(=O)[O-],\\ OC(=O)C1=CC(=[C]C(=C1)Oc1cccc(c1C(=O)[O-])C(=O)[O-])C(=O)O,\\ OC(=O)c1cc(cc(c1)C(=O)O)Oc1cccc(c1C(=O)[O-])C(=O)[O-], nodes [Ba][OH2][Ba], topology pts \\
output: 4.460577901 \\
\begin{figure*}[h!]
  \centering
  \includegraphics[width=0.5\textwidth]{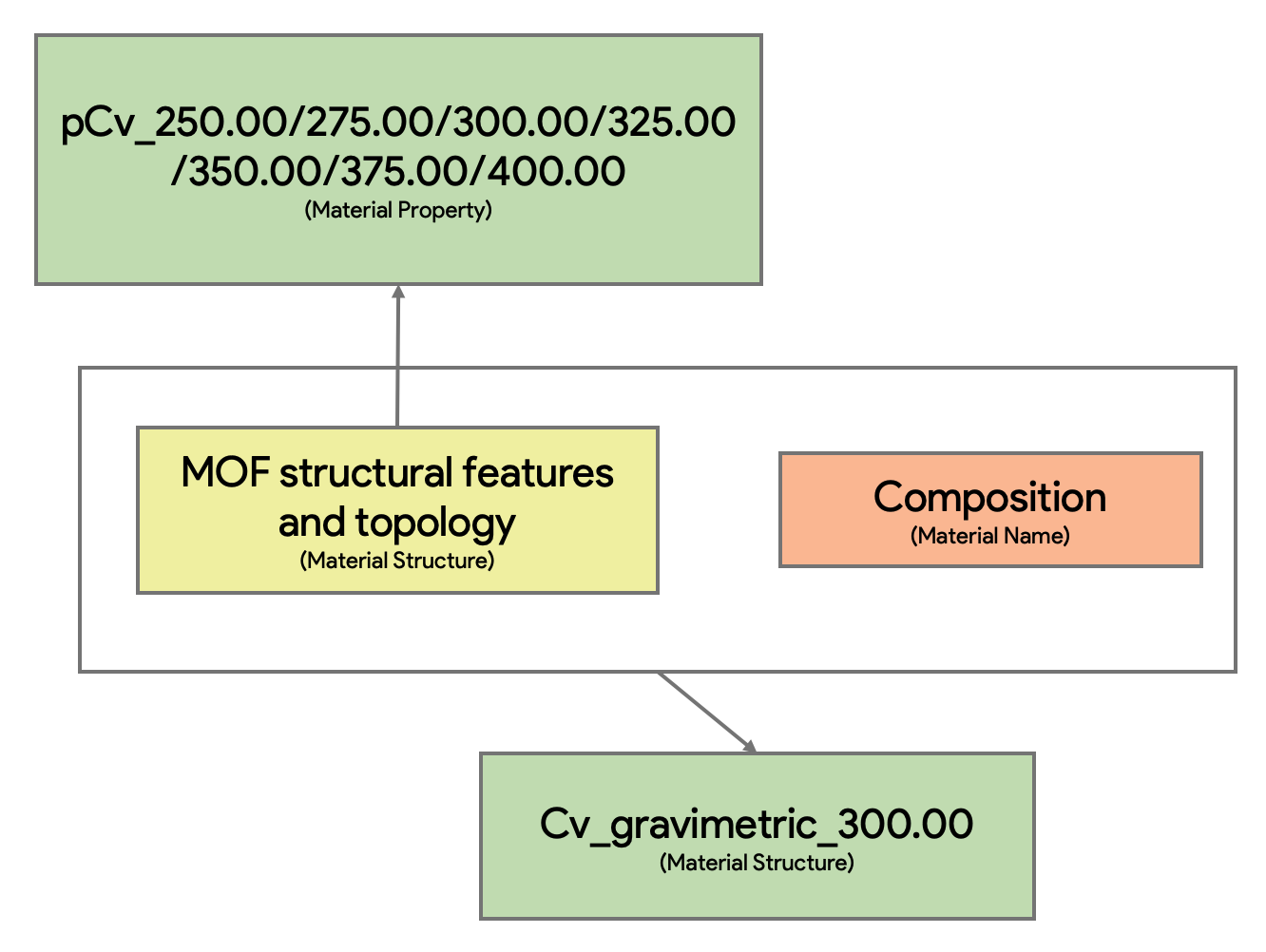}
  \caption{MoosaviCp task instruction generation structure,
A solid line delineates a one-to-one correspondence}
  \label{fig:pipeline}
\end{figure*} \\
\section{MDP classification task}
\textbf{Pei}: \\
instruction: What is phase of given alloy? \\
input: Ni0.9V0.1 \\
output: fcc \\
\textbf{WaterStability}: \\
instruction: How is the water stabilityof given stability at normal measurement conditions? \\
input: meso-tetra(4-carboxyl-phenyl)porphyrin(AlOH)2 \\
output: It has high water stability with medium confidence. \\
\begin{figure*}[h!]
  \centering
  \includegraphics[width=0.5\textwidth]{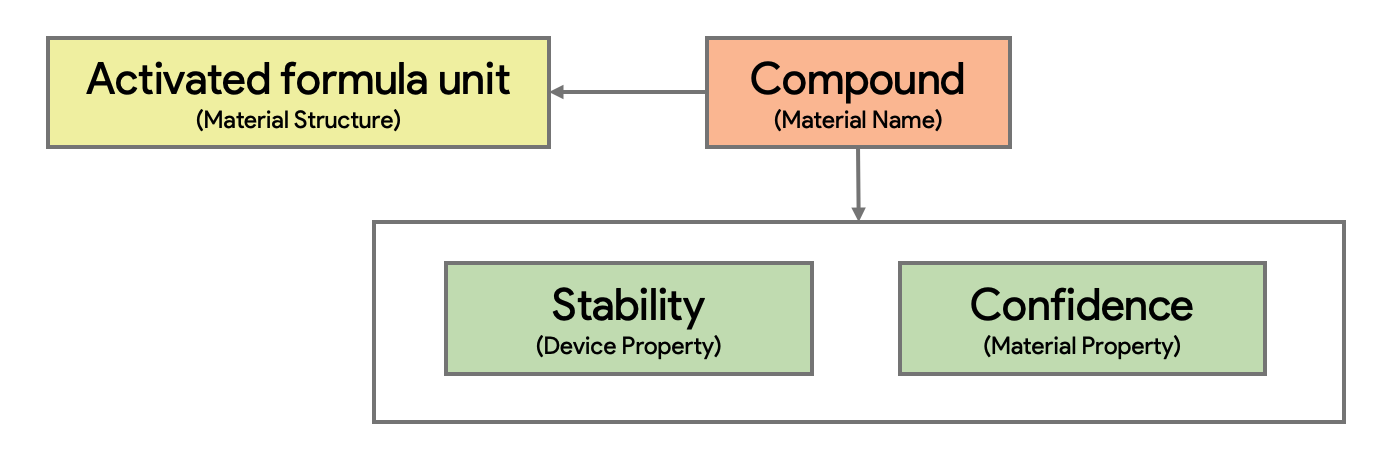}
  \caption{WaterStability task instruction generation structure,
A solid line delineates a one-to-one correspondence}
  \label{fig:pipeline}
\end{figure*} \\
\textbf{matbench is metal}: \\
instruction: Is given composition metal? \\
input: In5AgTe8 \\
output: No, In5AgTe8 is not metal. \\
\textbf{Matbench\_glass}: \\
instruction: Does given composition have glass formation ability? \\
input: Zr5(CrFe14)3 \\
output: Yes, Zr5(CrFe14)3 has glass formation ability. \\
\section{MDP inverse design task}
\textbf{OPV}: \\
instruction: Design a donor with acceptor PC61BM, PCE=0 for a organic solar cell \\
output: COC(=O)c1cc2csc(-c3cccs3)c2s1 \\

\section{Summary of FAIR Datasets}

\begin{table}[!htb]
%\footnotesize
  \centering
  %\caption{{Summary of FAIR Datasets}}
      \caption{Details of FAIR datasets and associated tasks they are used to derive instructions, (R for regression; D for inverse design; C for classification)}
    \label{tb:fair}
    \begin{tabular}{c|p{9.5cm}|c}
    \hline
    \textbf{Datasets}   & \textbf{Description}  & \textbf{Task}                 \\ \hline \hline
      
    \texttt{ChEMBL \cite{Gaulton2012}} & This dataset is sourced from a curated database of bioactive molecules with drug-like properties, focusing on the lipophilicity of 1899 molecular compounds in pharmacokinetics. The water-octanol partition coefficient (logD) is used to describe lipophilicity. & R
    \\ \hline
    
    \texttt{HybriD3 \cite{Qin2021}} & TThe HybriD materials database serves as a robust repository encompassing both experimental and computational data related to crystalline organic-inorganic compounds, with a predominant focus on the perovskite model. The data from 1518 materials are used in this work.  & R\&D  \\ \hline  

    \texttt{ESOL \cite{Delaney2004}} & This dataset is a compilation of measured aqueous solubility (LogS) values, a crucial factor in drug discovery. The dataset comprises 927 molecular compounds originally used for ESOL - estimated solubility.  & R  \\ \hline 

    \texttt{EPVRI \cite{Almora2023}} & The dataset is from the Emerging Photovoltaics Reports Initiative (EPVRI), which is an international academic platform in the field of emerging photovoltaic materials, including but not limited to organic, perovskite, and dye-sensitized solar cells. The dataset has the performance of 1177 solar cells  & R  \\ \hline
    
    \texttt{Magnet \cite{Sakurai2020}} & The database has a total of 3826 entries of magnetic materials, including 300+ Fe-based rare-earth-free magnets discovered through adaptive genetic algorithm (AGA) searches. 140 material results are used in this work. & R\\ \hline
    
    \texttt{Matbench\_expt\_gap \cite{Zhuo2018}} & This dataset is retrieved from Zhuo et al.'s work, containing data on experimental band gaps and DFT calculated zero band gaps for 4604 compounds. & R\\ \hline
    
    \texttt{Matbench\_glass \cite{Kawazoe1997}} & This dataset is retrieved from a volume of the Landolt– Börnstein collection 'Nonequilibrium phase diagrams of ternary amorphous alloys', containing data on full bulk metallic glass formation ability for 5680 chemical formulas.
    & C \\ \hline
    
    \texttt{Matbench\_is\_metal \cite{Dunn2020}} & This dataset is retrieved from Zhuo et al.'s work, containing data on classifying metallicity from composition for 4921 chemical formulas. & C \\ \hline
    
    \texttt{Matbench\_steels \cite{Dunn2020}} & This dataset is retrieved from Citrine informatics, containing data on steel yield strengths from composition for 312 chemical formulas & C  \\ \hline
    
    \texttt{MoosaviDiversity \cite{Moosavi2020} } & A diverse set of structures based on the chemical and geometric descriptors & R\\ \hline
    
    \texttt{MoosaviCp \cite{Moosavi2020}} & Dataset for predicting the heat capacity of materials based on density functional theory simulations & R \\ \hline
    
    \texttt{OPV \cite{Nagasawa2018}} & Experimental parameters including PCE, molecular weight, and electronic properties are manually collected from the literature and subjected to machine learning with digitized chemical structures & R \\ \hline
    
    \texttt{Pei \cite{Pei2020}} & The dataset is from a thermoelectric material review paper published in 2013, which abstracts information from over 100 publications. We used data of 1028 materials from this database. & C \\ \hline

    \texttt{PolarM \cite{Hickox-Young2023}} & Polar Metals Materials Database contains polar compounds exhibiting metallic conductivity as defined in the work of Hickox-Young et al. There are 191 materials in the dataset. & R  \\ \hline

    \texttt{TEM13 \cite{Sakurai2020}} & The dataset is from the work of Pei et al., originally used for machine learning training. It comprises information on 1252 alloys and their respective phases. & C \\ \hline
    
    \texttt{WaterStability \cite{Batra2020}} &  Consists of water stabilities for over 200 MOFs (metal–organic frameworks), alongside a comprehensive set of chemical features encompassing the metal node, organic ligand, and metal-ligand molar ratios. & C \\ \hline 
    
    \end{tabular}

\end{table}

\end{document}